\documentclass[letterpaper]{article} 
\usepackage{aaai23}  
\usepackage{times}  
\usepackage{helvet}  
\usepackage{courier}  
\usepackage[hyphens]{url}  
\usepackage{graphicx} 
\usepackage{natbib}  
\usepackage{caption} 
\usepackage{algorithm}
\usepackage{algorithmic}

\usepackage{graphicx}
\usepackage{amsfonts}
\usepackage{multirow}
\usepackage{array}
\usepackage{amsmath}

\usepackage{newfloat}
\usepackage{listings}
\DeclareCaptionStyle{ruled}{labelfont=normalfont,labelsep=colon,strut=off} 
\floatstyle{ruled}
\newfloat{listing}{tb}{lst}{}
\floatname{listing}{Listing}
\title{READ: Aggregating Reconstruction Error into Out-of-Distribution Detection}
\author {
    Wenyu Jiang,
    Yuxin Ge,
    Hao Cheng,
    Mingcai Chen,
    Shuai Feng,
    Chongjun Wang
}
\affiliations {
    State Key Laboratory for Novel Software Technology\\
    Nanjing University, Nanjing 210023, China \\
    \{lygjwy, yuxinge, chengh, chenmc, shuaifeng\}@smail.nju.edu.cn, chjwang@nju.edu.cn
}

\begin{document}

\maketitle

\begin{abstract}
Detecting out-of-distribution (OOD) samples is crucial to the safe deployment of a classifier in the real world. However, deep neural networks are known to be overconfident for abnormal data. Existing works directly design score function by mining the inconsistency from classifier for in-distribution (ID) and OOD. In this paper, we further complement this inconsistency with reconstruction error, based on the assumption that an autoencoder trained on ID data can not reconstruct OOD as well as ID. We propose a novel method, READ (\textbf{R}econstruction \textbf{E}rror \textbf{A}ggregated \textbf{D}etector), to unify inconsistencies from classifier and autoencoder. Specifically, the reconstruction error of raw pixels is transformed to latent space of classifier. We show that the transformed reconstruction error bridges the semantic gap and inherits detection performance from the original. Moreover, we propose an adjustment strategy to alleviate the overconfidence problem of autoencoder according to a fine-grained characterization of OOD data. Under two scenarios of pre-training and retraining, we respectively present two variants of our method, namely READ-MD (\textbf{M}ahalanobis \textbf{D}istance) only based on pre-trained classifier and READ-ED (\textbf{E}uclidean \textbf{D}istance) which retrains the classifier. Our methods do not require access to test time OOD data for fine-tuning hyperparameters. Finally, we demonstrate the effectiveness of the proposed methods through extensive comparisons with state-of-the-art OOD detection algorithms. On a CIFAR-10 pre-trained WideResNet, our method reduces the average FPR@95TPR by up to 9.8\% compared with previous state-of-the-art.
\end{abstract}

\section{Introduction}
Deep neural networks (DNNs) have attained high accuracy in image classification task \cite{zagoruyko2016wide}. However, the classifier often fails silently by providing overconfident prediction for input that belongs to a distribution different from the in-distribution (ID) of training data. Therefore, it is necessary to detect those out-of-distribution (OOD) samples for the deployment of classifier in safety-critical applications, such as autonomous driving and medical diagnosis.

For detecting OOD samples, the baseline method \cite{hendrycks2016baseline} utilizes the maximum value of posterior distribution from the pre-trained softmax classifier. They find that ID data tends to have greater prediction probabilities than OOD data. By temperature scaling and input perturbation, ODIN \cite{liang2017enhancing} improves the baseline method. However, it has been observed that softmax classifier can produce high confidence prediction for inputs far away from the training data \cite{hendrycks2016baseline,Nguyen_2015_CVPR}. The rationale is that the softmax classifier can have a label-overfitted output space \cite{Lee2018ASU,liu2020energy}. Instead of using the softmax outputs for OOD detection, Maha \cite{Lee2018ASU} assumes that pre-trained features of test data can be fitted well by a class-conditional Gaussian distribution and defines the confidence score using the Mahalanobis distance with respect to the closest class-conditional distribution in feature spaces. From probabilistic perspective of decomposing confidence, G-ODIN \cite{hsu2020generalized} uses a dividend/divisor structure for classifier. Then, the distance of input to the closest class is calculated with penultimate layer output of classifier to detect OOD samples.

The above methods are based on the observation that OOD data should be relatively far away from the ID classes. In this paper, we further complement the discrepancy of distance to the closest class in latent space. Based on the assumption that test data from the distribution same as training data can be better reconstructed than other distributions, we propose a reconstruction error aggregated detector (READ). The extracted representations by autoencoder are enforced to contain important regularities of the ID data. However, OOD inputs are poorly reconstructed from the resulting representations due to the irregular patterns. Our high-level idea is to mine the discrepancy of ID and OOD from classifier and autoencoder. To unify both discrepancies, i.e., the distance to the closest class and reconstruction error, we transform raw pixels reconstruction error to the latent space of classifier. Overall, the transformed reconstruction error exhibits competitive OOD detection performance compared with the raw pixels. Based on the same reconstruction error assumption, Gong et al. \cite{gong2019memorizing} and Zhang et al. \cite{zhang2021label} incorporate memory module to autoencoder and directly use raw pixels reconstruction error to detect OOD samples. However, they find that this assumption does not always hold and the autoencoder can reconstruct specific OOD data well with low reconstruction error. Similar overconfident phenomenon for flow-based deep generative models is reported in \cite{choi2018waic,nalisnick2018deep}. For transformed reconstruction error, we observe that the same overconfidence problem. In order to alleviate this problem, we further propose a fine-grained characterization of OOD based on \cite{hsu2020generalized}. Then, we introduce a coefficient to adjust transformed reconstruction error according to the data types. Empirical result shows that adjustment coefficient alleviates the overconfidence problem. Under two scenarios of pre-training and retraining, we propose corresponding variants of READ, namely READ-MD (\textbf{M}ahalanobis \textbf{D}istance) only based on pre-trained classifier and READ-ED (\textbf{E}uclidean \textbf{D}istance) which retrains the modified classifier.

The complete illustration of our method is presented in Figure ~\ref{fig1}. Through extensive and comprehensive evaluations on common OOD detection benchmarks, both of our methods, READ-MD and READ-ED, achieve state-of-the-art performance compared with previously best methods under corresponding scenarios. In ablation studies, we also demonstrate the effectiveness of the proposed transformed reconstruction error and adjustment coefficient. Note that the choice of \textbf{hyperparameters does not rely on test time OOD data} and \textbf{no auxiliary OOD samples are provided at training time.}

Our main contributions are summarized as follows:
\begin{itemize}
    \item We propose a novel reconstruction error aggregated detector (READ) and its two variants, READ-MD and READ-ED, which combine the distance to the closest class and reconstruction error in the latent space of classifier. 
    \item Against the overconfidence of transformed reconstruction error, we explain and alleviate this problem by a fine-grained characterization of OOD data and an image complexity based adjustment coefficient.
    \item We conduct comprehensive analysis with experiments under both scenarios to demonstrate the effectiveness of the proposed methods.
\end{itemize}

\section{Related Work}

\subsubsection{Out-of-distribution Detection for Discriminative Models.}
Given pre-trained classifier, a baseline method \cite{hendrycks2016baseline} utilizing maximum softmax probability (MSP) is proposed based on the observation that ID samples tend to have greater prediction probabilities than OOD samples. However, the MSP score for OOD input is proven to be arbitrarily high for neural networks with ReLU activation \cite{Hein_2019_CVPR}. Liang et al. \cite{liang2017enhancing} improve the baseline with temperature scaling and input perturbation techniques, and further enlarge the gap between ID and OOD data. Instead of deriving score function from label-overfitted output space, Lee et al. \cite{Lee2018ASU} and Sastry et al. \cite{sastry2020detecting} design confidence score in feature spaces of the pre-trained classifier. Liu et al. \cite{liu2020energy} propose energy score which can be easily derived from the logit output of the pre-trained classifier and demonstrate superiority over softmax score both empirically and theoretically. Sun et al. \cite{sun2021react} show that a simple activation rectification strategy termed ReAct can significantly improve OOD detection performance. Recent work by Huang et al. \cite{huang2021importance} proposes a score function named GradNorm from the gradient space. GradNorm utilizes the vector norm of gradients, backpropogated from the KL divergence between the softmax output and a uniform probability distribution.

Loosening the restriction on retraining, G-ODIN \cite{hsu2020generalized} modifies the classifying head with a dividend/divisor structure for decomposing confidence of predicted class probabilities. Moreover, a modified input perturbation strategy is proposed to remove the unrealistic requirement of previous methods \cite{liang2017enhancing,Lee2018ASU} that the choice of hyperparameters depends on test time OOD data. In \cite{tack2020csi,sehwag2021ssd}, self-supervised learning is used to learn better visual representations for OOD detection. In our work, we further complement the discrepancy of ID and OOD from discriminative models with reconstruction error.

\subsubsection{Out-of-distribution Detection for Generative Models.}
There are several works that detect OOD samples with generative models. The input data is defined as OOD if it lies in the low-density regions. However, as shown in \cite{choi2018waic,nalisnick2018deep}, flow-based generative models \cite{kingma2013auto,van2016conditional,rezende2014stochastic} can assign a high likelihood to OOD data. This problem is addressed by considering a likelihood ratio \cite{ren2019likelihood}, taking the input complexity into account \cite{serra2019input}, and likelihood regret \cite{xiao2020likelihood}. For autoencoder, the specific OOD inputs can be reconstructed well demonstrated by  \cite{denouden2018improving,gong2019memorizing,zhang2021label}. In contrast, we transform the reconstruction error to latent space of classifier as a supplement and propose an adjustment coefficient to alleviate the overconfidence problem.

\subsubsection{Characterization of Out-of-distribution Data.}
According to \cite{hsu2020generalized}, OOD data can be described from covariate shift and semantic shift perspectives. Works in OOD detection task focus on detection of semantic shift OOD inputs. We further characterize the semantic shift OOD data based on image complexity \cite{lin2021mood,serra2019input} for explaining and alleviating the varying detection performance of autoencoder. Some works propose a fine-grained characteization of covariate shift OOD data \cite{hendrycks2019benchmarking,ovadia2019can} for evaluation of model, including corruption-shift for robustness and domain-shift for domain generalization performance. Note that our work is dedicated to finding new concepts at inference time.

\section{Method}
\begin{figure*}[htb]
\includegraphics[width=\textwidth]{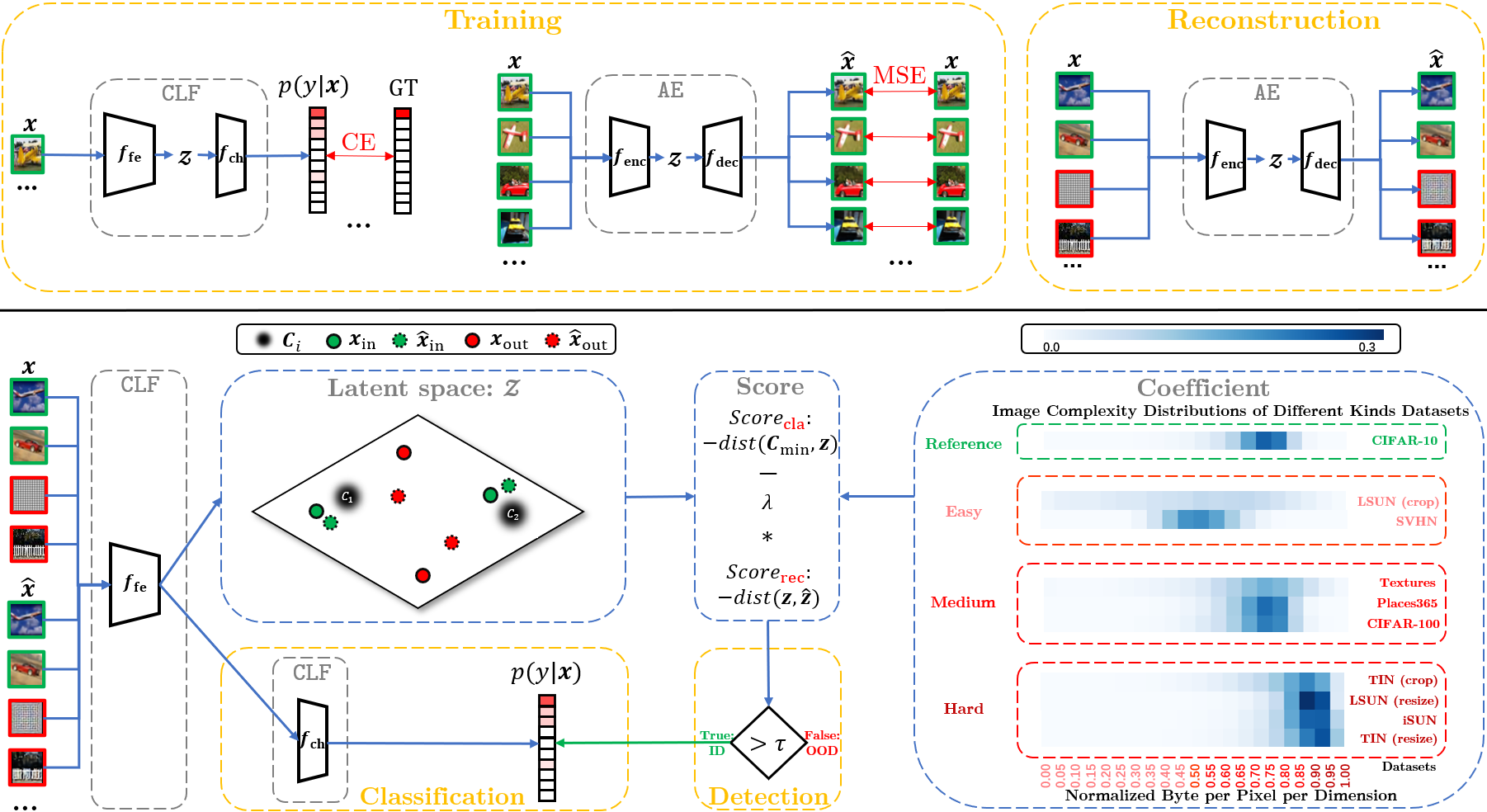}
\caption{Illustration of the proposed reconstruction error aggregated detector (READ). \emph{Top}: Architecture of the proposed detector and preparatory phase (training and reconstruction). \emph{Bottom}: Overview of the OOD detection and classification procedure at detection phase (detection and classification).}
\label{fig1}
\end{figure*}

Our method is illustrated in Figure \ref{fig1}. In this section, we first formalize the out-of-distribution detection problem. Secondly, we present the overall concept of our algorithm. Then, we detail the actual training process. To avoid confusion, the score function is illustrated separately under both scenarios. We explain the reconstruction error adjustment coefficient based on image complexity. Finally, we present the OOD detection and classification procedure at inference time.

\subsection{Problem Statement}
In this paper, we consider the OOD detection problem under setting of multi-category image classification. Let $\mathcal{X} = \mathbb{R}^d$ denote the input space and $\mathcal{Y} = {\{1,...,K\}}$ denote the corresponding label space. We have access to the labeled training set $\mathcal{D}^{\mathrm{train}}_{\mathrm{in}} = {\{(\boldsymbol{x}_{i}, y_{i})\}}^{n}_{i=1}$, drawn \emph{i.i.d} from the joint data distribution $\mathbb{P_{\mathcal{X} \times \mathcal{Y}}}$. Let $f_{\theta} : \mathcal{X} \mapsto \mathbb{R}^{|\mathcal{Y}|}$ denote a neural network for the classification task, which predicts the label of an input sample. Furthermore, we denote the marginal probability distribution on $\mathcal{X}$ by $\mathbb{P}_{\mathcal{X}}^{\mathrm{in}}$, which represents the distribution of in-distribution data. At inference time, the classifier $f$ will encounter a different distribution $\mathbb{P}_{\mathcal{X}^{\mathrm{out}}}$ over $\mathcal{X}$ of out-of-distribution data. Out-of-distribution detection aims to design a binary function estimator,
$$
g(x)=
\begin{cases}
1,      & \mathrm{if} \ \boldsymbol{x} \sim \mathbb{P}_{\mathcal{X}}^{\mathrm{in}} \\
0,      & \mathrm{if} \ \boldsymbol{x} \sim \mathbb{P}_{\mathcal{X}}^{\mathrm{out}}
\end{cases}
$$
that classifies whether a test-time sample $\boldsymbol{x} \in \mathcal{X}$ is generated from $\mathbb{P}_{\mathcal{X}}^{\mathrm{in}}$ or $\mathbb{P}_{\mathcal{X}}^{\mathrm{out}}$. In practice, the $\mathbb{P}_{\mathcal{X}}^{\mathrm{out}}$ is often defined by an irrelevant distribution with non-overlapping labels with regard to in-distribution data. Hence, the classifier $f$ should not predict OOD data.

\subsection{Overall Concept}
Based on the reconstruction error assumption, we introduce autoencoder into out-of-distribution detection. As shown in the Training part of Figure \ref{fig1}, the network architecture of our method consists of two components: (a) a classifier (\texttt{CLF}) containing feature extractor $f_{\mathrm{fe}}$ for learning latent representations $\boldsymbol{z}$ with parameters $\theta_{\mathrm{fe}}$ and classifying head $f_{\mathrm{ch}}$ with parameters $\theta_{\mathrm{ch}}$ which takes $\boldsymbol{z}$ and classify them to known classes. The notation $p(y|\boldsymbol{x})$ denotes the prediction posterior distribution for input $\boldsymbol{x}$. (b) an autoencoder (\texttt{AE}), including encoder $f_{\mathrm{enc}}$ to compress high-dimensional data features with parameters $\theta_{\mathrm{enc}}$ and deconder $f_{\mathrm{dec}}$ recreating $\boldsymbol{x}$ denoted by $\hat{\boldsymbol{x}}$ from the latent representation $\boldsymbol{z}$ with parameters $\theta_{\mathrm{dec}}$. Different from works \cite{oza2019c2ae,zhang2021label} integrating classifier and autoencoder in one hybrid model simultaneously and utilizing raw pixels reconstruction error as score function, the \texttt{CLF} and \texttt{AE} modules in our method are independent of each other. Furthermore, we transform the reconstruction error to \texttt{CLF} latent space instead of pixels space for further aggregation. For one thing, the transformed reconstruction error bridges the semantic gap, and for another, we empirically show its superior detection performance. Lee at el. \cite{Lee2018ASU} also remark that OOD samples can be characterized better by latent embeddings of \texttt{CLF}, rather than the ``label-overfitted'' output space. These take us to transform reconstruction error to the latent space of \texttt{CLF}.

After transforming the reconstruction error, we combine it with the distance of the input $\boldsymbol{x}$ to the closest category $\mathcal{C}_i$ in the latent space of \texttt{CLF} since the OOD inputs should be relatively far away from the ID classes. We illustrate our idea in latent space part of Figure \ref{fig1}. Obviously, the combination of transformed reconstruction error and distance to the closest class in latent space brings better separability of ID and OOD samples. We proceed with detailing the training procedure, OOD detection and classification procedures.

\subsection{Training}
As described above, our architecture contains two independent components: \texttt{CLF} and \texttt{AE}. In training stage, we need to train \texttt{CLF} to classify ID samples correctly and train \texttt{AE} to reconstruct original inputs. Specifically, our \texttt{CLF} is trained to optimize parameters $\theta_{\mathrm{fe}}$ and $\theta_{\mathrm{ch}}$ by minimizing the following cross-entropy loss function:
\begin{equation}
\mathcal{L}_{\mathrm{CLF}} = \mathbb{E}_{(\boldsymbol{x},y)\sim\mathcal{D}^{\mathrm{train}}_{\mathrm{in}}}[-\log F_y(\boldsymbol{x})]
\end{equation}
where $F_y(\boldsymbol{x})$ is the softmax output of \texttt{CLF}. For the \texttt{AE}, the parameters $\theta_{\mathrm{enc}}$ and $\theta_{\mathrm{dec}}$ are updated. The training mean-squared error loss function is as follows:
\begin{equation}
\mathcal{L}_{\mathrm{AE}} = \mathbb{E}_{\boldsymbol{x}\sim\mathcal{X}^{\mathrm{train}}_{\mathrm{in}}}[\|\boldsymbol{x} - {\hat{\boldsymbol{x}}}\|^{\mathrm{2}}_{\mathrm{2}}]
\end{equation}
where $\hat{\boldsymbol{x}}$ is the reconstruction output of \texttt{AE} and $\mathcal{X}^{\mathrm{train}}_{\mathrm{in}}$ is the ID training data without labels. The complete training procedures are illustrated in Training part of Figure \ref{fig1}.

\subsection{Transformed Reconstruction Error}
To measure the distance of test input $\boldsymbol{x}$ to the closest category $\mathcal{C}_i$ and reconstruction error in latent space $\mathcal{Z}$, we first need to model classes by ID training data. Considering the limitations of \texttt{CLF} retraining in practical problems, we adopt two different class modeling strategies with corresponding distance metrics, namely Mahalanobis distance and Euclidean distance.

\subsubsection{Pre-training Scenario.}
Given the pre-trained \texttt{CLF} without subsequent retraining, we use the same class modeling method as \cite{Lee2018ASU}. We define $K$ class-conditional distributions with a tied covariance $\mathbf{\Sigma}$: $p(f_{\mathrm{fe}}(\boldsymbol{x})|y=i) = \mathcal{N}(f_{\mathrm{fe}}(\boldsymbol{x})|\mathbf{\mu}_i,\mathbf{\Sigma})$, where $\mathbf{\mu}_i$ is the mean of multivariate Gaussian distribution of class $i \in \{1,...K\}$, assuming that the class-conditional distribution of \texttt{CLF} latent representations follows multivariate Gaussian distribution. Then, the empirical class mean and covariance of training data are computed to estimate the parameters of the class-conditional distribution as follows:
\begin{equation}
\hat{\mathbf{\mu}}_i = \mathbb{E}_{\boldsymbol{x}\sim\mathcal{X}^{\mathrm{train}}_{\mathrm{in}},y=i}[f_{\mathrm{fe}}(\boldsymbol{x})]
\end{equation}
\begin{equation}
\hat{\mathbf{\Sigma}} = \mathbb{E}_{(\boldsymbol{x},y)\sim\mathcal{D}^{\mathrm{train}}_{\mathrm{in}}}[(f_{\mathrm{fe}}(\boldsymbol{x})-\hat{\mathbf{\mu}}_y)({f_{\mathrm{fe}}(\boldsymbol{x})-\hat{\mathbf{\mu}}_y})^\mathsf{T}]
\end{equation}
After modeling the ID classes with multivariate Gaussian distributions, we measure distance between test input $\boldsymbol{x}$ and the closest class-conditional distribution by Mahalanobis distance, i.e.,
\begin{equation}
\mathit{Score}_{\mathrm{cla}} = - \min_i (f_{\mathrm{fe}}(\boldsymbol{x})-\hat{\mathbf{\mu}}_i)^\mathsf{T}{\hat{\mathbf{\Sigma}}}^{-1}(f_{\mathrm{fe}}(\boldsymbol{x})-\hat{\mathbf{\mu}}_i))
\end{equation}
Accordingly, the definition of reconstruction error between original data $\boldsymbol{x}$ and reconstructed data $\hat{\boldsymbol{x}}$ in latent space of \texttt{CLF} is presented below:
\begin{equation}
\mathit{Score}_{\mathrm{rec}} = -  ((f_{\mathrm{fe}}(\boldsymbol{x})-f_{\mathrm{fe}}(\hat{\boldsymbol{x}}))^\mathsf{T}{\hat{\mathbf{\Sigma}}}^{-1}(f_{\mathrm{fe}}(\boldsymbol{x})-f_{\mathrm{fe}}(\hat{\boldsymbol{x}}))
\end{equation}

\subsubsection{Retraining Scenario.}
Loosening restriction on the retraining of \texttt{CLF},  Hsu at el. \cite{hsu2020generalized} change the original \texttt{CLF}'s $f_{\mathrm{ch}}$ from fully connected layer to a dividend/divisor structure with a novel perspective of decomposed confidence. Inspired by this, we modify the $f_{\mathrm{ch}}$ based on their work as follows:
\begin{equation}
f_{\mathrm{ch}_i}(\boldsymbol{z}) = \frac{h_{i}(\boldsymbol{z})}{g(\boldsymbol{z})} = \frac{-\|\boldsymbol{z} - {\mathbf{\omega}_i}\|^{\mathrm{2}}_{\mathrm{2}}}{\sigma(BN(\mathbf{\omega}_{\mathrm{g}}\boldsymbol{z} + b_{\mathrm{g}}))}
\end{equation}
where $\boldsymbol{z}$ is the output of $f_{\mathrm{fe}}$ and $BN$ denotes the batch normalization layer. In proportion, the ID class centers are fitted by learnable parameters of classifier, i.e., $w_i$. Moreover, the distance of input $\boldsymbol{x}$ to the closest class center and transformed reconstruction error in latent space of \texttt{CLF} are defined using Euclidean distance as follows:
\begin{equation}
\mathit{Score}_{\mathrm{cla}} = - \min_i (\|\boldsymbol{z} - {\mathbf{\omega}_i}\|^{\mathrm{2}}_{\mathrm{2}})
\end{equation}
\begin{equation}
\mathit{Score}_{\mathrm{rec}} = - (\|\boldsymbol{z} - \hat{\boldsymbol{z}}\|^{\mathrm{2}}_{\mathrm{2}})
\end{equation}
where $\hat{\boldsymbol{z}}$ is the output of $f_{\mathrm{fe}}$ when the input is $\hat{\boldsymbol{x}}$. Note that we do not use auxiliary OOD training data in both scenarios.

\subsection{Adjustment Coefficient}

\begin{figure*}
\includegraphics[width=\textwidth]{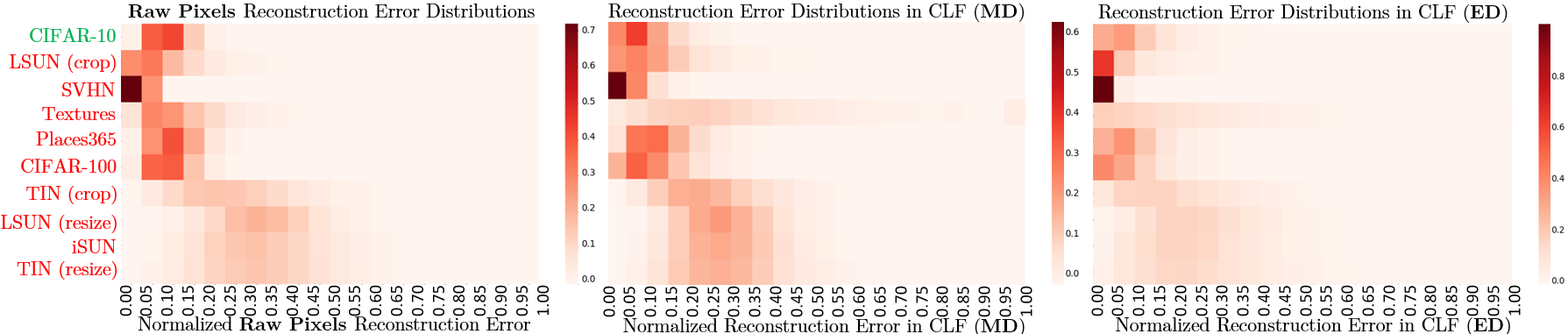}
\caption{The reconstruction error distributions in different forms (CIFAR-10 as ID).}
\label{fig2}
\end{figure*}

Although the transformed reconstruction error brings superior discrimination, we observe that the detection performance is inconsistent across various OOD datasets, as shown in Figure \ref{fig2}. Under different metrics, we find the same three distribution patterns of OODs when CIFAR-10 is taken as ID reference. Specifically, the distribution is skewed to the smaller reconstruction error for ``easy'' OODs which contain simple objects or constant pixels since simpler representations are required for their description. For ``medium'' OODs which have the covariate same as ID, the distributions are similar, i.e., it is indistinguishable of inputs by reconstruction error. In general, the fact that \textbf{\texttt{AE} trained on ID data can reconstruct the ``easy'' and ``medium'' OOD data well with low reconstruction error} poses a challenge to OOD detection. Explaining from the multi-category learning process of \texttt{AE}, the diversity of training data increases the difficulty of ID reconstruction compared to the single class. Similarly, the overconfident phenomenon is also reported in \cite{choi2018waic,denouden2018improving,gong2019memorizing,nalisnick2018deep,zhang2021label}. Lastly, the reconstruction error distribution of ``hard'' OODs which contain richer contents and diverse pixels compared to ID is skewed to the right side as expected, and this is consistent with the reconstruction error assumption. The rationale is that the learned representations by \texttt{AE} are enforced to learn important regularities of the ID data to minimize reconstruction errors. Hence, OOD data are poorly reconstructed from the resulting representations.

To sum up, the reconstruction error assumption does not always hold for different kinds of OOD data and this conclusion is applicable to different reconstruction error forms.

In order to alleviate the above issue, we firstly propose a fine-grained characterization of OODs based on  \cite{hsu2020generalized} to deal with different kinds of reconstruction error distribution patterns. Concretely, we adopt a complexity score as a proxy measurement to quantify the ``easiness'' of OODs by off-the-shelf lossless image compression algorithm \cite{lin2021mood,serra2019input}. As shown in Coefficient part of Figure \ref{fig1}, considering the essence of OOD detection is to find novel concepts at inference time, we further characterize semantic shift OOD to three kinds by the lower and upper complexity bound of ID training data removing extreme samples, i.e., the easiest and hardest top 5\%. For example, the SVHN test images which have smaller complexities than lower complexity bound are categorized to easy OOD. Note that the test images whose complexities lie within the range of lower and upper complexity bound can be medium OOD or ID. Then, according to the type of inputs, we adjust and re-scale transformed reconstruction error with coefficient $\lambda$. We simply set $\lambda$ to 0.5 for ID and keep the original reconstruction error for easy and hard OODs. Hence, the gap between easy, hard OODs and ID is enlarged. Besides, the finer characterization of OODs can serve as a principle to design equitable benchmark protocol. We notice that the experimental setup of dividing a multi-class dataset into ID and OOD adopted by \cite{ahmed2020detecting} and many OSR works is inadequate to evaluate OOD detector because they only consider medium OODs.

\subsection{Inference}
At inference time, an input image $\boldsymbol{x}$ and corresponding reconstruction $\hat{\boldsymbol{x}}$ are forward propagated through $f_{\mathrm{fe}}$. For classification, the latent representation $\boldsymbol{z}$ is further propagated through $f_{\mathrm{ch}}$. For OOD detection, we use two metrics under both scenarios to compute $Score_{\mathrm{cla}}$ and $Score_{\mathrm{rec}}$ in latent space, i.e., Mahalanobis distance and Euclidean distance. We call the two variants of our method READ-MD and READ-ED. Then, the $Score_{\mathrm{rec}}$ is adjusted by coefficient $\lambda$ based on image complexity. The final score is as follows:
\begin{equation}
Score = -Score_{\mathrm{cla}} - \lambda * Score_{\mathrm{rec}}
\end{equation}
When the $Score$ is above a detection threshold $\tau$, we assign the test input $\boldsymbol{x}$ as an ID sample. 
The above procedure is illustrated in the lower part of Figure \ref{fig1}.

Additionally, we adopt input perturbation strategy proposed in \cite{liang2017enhancing}. They find that input perturbation brings larger gain on $Score$ for ID samples. We modify original strategy by perturbing over $Score_{\mathrm{cla}} + Score_{\mathrm{rec}}$. In detail, the perturbation of input $\boldsymbol{x}$ is given by:
\begin{equation}
\tilde{\boldsymbol{x}} = \boldsymbol{x} - \epsilon * \mathrm{sign}(-\nabla_{\boldsymbol{x}}(Score_{\mathrm{cla}}(\boldsymbol{x})+Score_{\mathrm{rec}}(\boldsymbol{x}, \hat{\boldsymbol{x}})))
\end{equation}
Then, $Score$ is recalculated with $\tilde{\boldsymbol{x}}$ and $\hat{\boldsymbol{x}}$ as described previously. Considering that test time OOD data is unavailable, the choice of hyperparameters depends on metric FPR@TPR95 of ID and synthesized OOD data from \cite{hendrycks2018deep}, including uniform noise and 7 kinds corrupted ID samples, i.e., arithmetic mean, geometric mean, jigsaw, speckle noised, pixel, RGB ghosted, and inverted.

\section{Experiments}
\begin{table*}[ht]
\begin{tabular}{>{\centering\arraybackslash}p{0.05\textwidth}>{\centering\arraybackslash}p{0.15\textwidth}>{\centering\arraybackslash}p{0.35\textwidth}>{\centering\arraybackslash}p{0.35\textwidth}}
\hline
\multirow{2}*{\textbf{ID}} & \multirow{2}*{\textbf{OOD}} & \textbf{FPR@95TPR} $\downarrow$ & \textbf{AUROC} $\uparrow$ \\
\cline{3-4}
& & \multicolumn{2}{c}{MSP\,/\,ODIN\,/\,Maha\,/\,Energy\,/\,READ-MD (ours)} \\
\hline
\multirow{10}*{\rotatebox[origin=c]{90}{CIFAR-10}} 
& SVHN &        48.3\,/\,33.2\,/\,15.3\,/\,35.4\,/\,\textbf{12.0}           & 91.9\,/\,92.0\,/\,97.0\,/\,91.1\,/\,\textbf{97.5} \\
& LSUN (c) &    42.4\,/\,29.7\,/\,31.6\,/\,\textbf{19.1}\,/\,28.3           & 93.6\,/\,92.8\,/\,94.1\,/\,\textbf{96.0}\,/\,94.9 \\
& Textures &    59.5\,/\,49.5\,/\,18.0\,/\,52.5\,/\,\textbf{10.3}           & 88.4\,/\,84.7\,/\,96.3\,/\,85.4\,/\,\textbf{98.0} \\
& Places365 &   60.5\,/\,57.7\,/\,74.2\,/\,\textbf{40.9}\,/\,75.5           & 88.1\,/\,84.3\,/\,80.3\,/\,\textbf{89.7}\,/\,80.7 \\
& CIFAR-100 &   62.9\,/\,60.7\,/\,71.8\,/\,\textbf{50.5}\,/\,76.5           & \textbf{87.8}\,/\,82.7\,/\,79.7\,/\,87.1\,/\,79.2 \\
& TIN (c) &     54.3\,/\,37.3\,/\,37.7\,/\,38.3\,/\,\textbf{19.9}           & 90.5\,/\,91.6\,/\,92.9\,/\,91.5\,/\,\textbf{96.5} \\
& LSUN (r) &    52.0\,/\,26.5\,/\,34.1\,/\,27.9\,/\,\textbf{9.4}            & 91.5\,/\,94.6\,/\,94.2\,/\,94.1\,/\,\textbf{98.3} \\
& TIN (r) &     60.8\,/\,39.1\,/\,34.1\,/\,46.5\,/\,\textbf{12.3}           & 88.2\,/\,91.3\,/\,93.5\,/\,89.0\,/\,\textbf{97.7} \\
& iSUN &        56.4\,/\,32.4\,/\,33.5\,/\,33.9\,/\,\textbf{12.5}           & 89.9\,/\,93.4\,/\,93.9\,/\,92.6\,/\,\textbf{97.6} \\
\cline{2-4}
& \textbf{average} &  55.2\,/\,40.7\,/\,38.9\,/\,38.3\,/\,\textbf{28.5}     & 90.0\,/\,89.7\,/\,91.3\,/\,90.7\,/\,\textbf{93.4} \\
\hline

\multirow{10}*{\rotatebox[origin=c]{90}{CIFAR-100}} 
& SVHN &        85.0\,/\,82.1\,/\,\textbf{58.0}\,/\,92.2\,/\,67.9           & 70.3\,/\,69.1\,/\,\textbf{85.3}\,/\,73.6\,/\,81.8 \\
& LSUN (c) &    79.0\,/\,66.8\,/\,63.5\,/\,75.4\,/\,\textbf{61.7}           & 77.6\,/\,81.2\,/\,82.0\,/\,83.1\,/\,\textbf{83.1} \\
& Textures &    83.1\,/\,78.8\,/\,36.9\,/\,78.0\,/\,\textbf{35.6}           & 73.4\,/\,72.9\,/\,90.9\,/\,76.0\,/\,\textbf{92.1} \\
& Places365 &   82.9\,/\,88.4\,/\,90.6\,/\,\textbf{81.3}\,/\,91.7           & 73.4\,/\,70.5\,/\,64.5\,/\,\textbf{75.4}\,/\,63.3 \\
& CIFAR-10 &    \textbf{81.8}\,/\,89.2\,/\,93.9\,/\,82.4\,/\,95.0           & 75.1\,/\,70.1\,/\,61.9\,/\,\textbf{77.2}\,/\,69.3 \\
& TIN (c) &     78.5\,/\,74.4\,/\,41.5\,/\,63.1\,/\,\textbf{29.8}           & 76.5\,/\,80.0\,/\,91.0\,/\,81.2\,/\,\textbf{93.6} \\
& LSUN (r) &    82.5\,/\,73.9\,/\,22.7\,/\,62.0\,/\,\textbf{10.9}           & 74.5\,/\,80.3\,/\,95.7\,/\,79.1\,/\,\textbf{97.6} \\
& TIN (r) &     82.3\,/\,71.6\,/\,25.3\,/\,63.5\,/\,\textbf{14.7}           & 73.7\,/\,80.2\,/\,94.8\,/\,77.5\,/\,\textbf{97.0} \\
& iSUN &        83.1\,/\,70.6\,/\,26.2\,/\,62.3\,/\,\textbf{15.5}           & 75.0\,/\,81.4\,/\,94.3\,/\,78.9\,/\,\textbf{96.3} \\
\cline{2-4}
& \textbf{average} &  82.0\,/\,77.3\,/\,51.0\,/\,73.4\,/\,\textbf{47.0}     & 74.4\,/\,76.2\,/\,84.5\,/\,78.0\,/\,\textbf{84.9} \\
\hline
\end{tabular}
\caption{Comparison with post-hoc methods. $\uparrow$ ($\downarrow$) indicates larger (smaller) values are better. \textbf{Bold} numbers are superior.}
\label{tab1}
\end{table*}

In this section, we describe our experimental setup and demonstrate the effectiveness of our proposed method on various benchmark setups. Also, we conduct extensive ablation studies to explore different aspects of our algorithm. We ran all experiments using PyTorch 1.4.0.

\subsection{Setup}
\subsubsection{In-distribution Datasets.}
CIFAR-10 (contains 10 classes) \cite{krizhevsky2009learning}, and CIFAR-100 (contains 100 classes) \cite{krizhevsky2009learning} datasets are used as in-distribution data. We use the standard split, training set for training deep neural networks for image classification and reconstruction, and test set for evaluation.
\subsubsection{Out-of-distribution Datasets.}
Considering the fine-grained characterization of OOD datasets, we use ten common benchmarks used in  \cite{liang2017enhancing,liu2020energy,tack2020csi} for the comprehensiveness and fairness of evaluation as OOD test datasets: SVHN \cite{netzer2011reading}, CIFAR-10, CIFAR-100, Textures \cite{cimpoi2014describing}, Places365 \cite{zhou2017places}, TinyImageNet (crop) \cite{deng2009imagenet}, TinyImageNet (resize) \cite{deng2009imagenet}, LSUN (crop) \cite{yu2015lsun}, LSUN (resize) \cite{yu2015lsun}, and iSUN \cite{xu2015turkergaze}.
In order to avoid overlapping with OOD validation data, we do not adopt uniform noise data. TinyImageNet (crop), TinyImageNet (resize), LSUN (crop), LSUN (resize), and iSUN are provided as a part of \cite{liang2017enhancing} code release.\footnote{https://github.com/facebookresearch/odin} Note that we preprocess cropped datasets with center clipping to remove the black border. We adopt officially original versions of the remaining datasets. For Places365, we use the same sampling as \cite{chen2021atom} for experimental results reproduction. The sampling list is publicly available at their code release.\footnote{https://github.com/jfc43/informative-outlier-mining} All images are down-sampled to 32 $\times$ 32.

\subsubsection{Networks and Training Details.}
We use WideResNet \cite{zagoruyko2016wide}, with depth 40, width 2 and dropout rate 0.3 as the classifier backbone. For READ-MD, we directly use the pre-trained classification model provided by \cite{liu2020energy} at their code release.\footnote{https://github.com/wetliu/energy\_ood} For READ-ED, we follow training details of \cite{hsu2020generalized}, the classifier is trained with batch size 128 for 200 epochs with weight decay 0.0005. The optimizer is SGD with momentum 0.9, and the learning rate starts with 0.1 and decreases by factor 0.1 at 50\% and 75\% of the training epochs. The weights in $h_{i}(\boldsymbol{x})$ of classifier are initialized with He-initialization \cite{he2015delving} and not applied with weight decay. As for reconstruction model, we design an vanilla autoencoder with symmetrical structure, using ResNet18 \cite{he2016deep} as encoder to deal with complex multi-class training data. The autoencoder is trained with batch size 128 for 2,000 epochs without weight decay. The optimizer is Adam with learning rate 0.001, betas 0.9 and 0.999. During training, we augment our training data with random flip and random cropping.

\subsubsection{Evaluation Metrics.}
We measure the following metrics: (1) the area under the receiver operating characteristic curve (AUROC); and (2) the false positive rate of OOD examples when true positive rate of ID data is at 95\% (FPR@95TPR).

\begin{table*}[htp]
\begin{tabular}{>{\centering\arraybackslash}p{0.05\textwidth}>{\centering\arraybackslash}p{0.15\textwidth}>{\centering\arraybackslash}p{0.35\textwidth}>{\centering\arraybackslash}p{0.35\textwidth}}
\hline
\multirow{2}*{\textbf{ID}} & \multirow{2}*{\textbf{OOD}} & \textbf{FPR@95TPR} $\downarrow$ & \textbf{AUROC} $\uparrow$ \\
\cline{3-4}
& & \multicolumn{2}{c}{G-ODIN-I\,/\,G-ODIN-C\,/\,G-ODIN-E\,/\,READ-ED (ours)} \\
\hline
\multirow{10}*{\rotatebox[origin=c]{90}{CIFAR-10}} 
& SVHN &            11.1\,/\,9.7\,/\,\textbf{8.3}\,/\,10.3          & 98.0\,/\,98.1\,/\,\textbf{98.2}\,/\,97.9 \\
& LSUN (c) &        6.1\,/\,11.0\,/\,3.1\,/\,\textbf{2.8}           & 98.9\,/\,97.9\,/\,99.3\,/\,\textbf{99.4} \\
& Textures &        26.6\,/\,22.0\,/\,19.3\,/\,\textbf{14.9}        & 94.9\,/\,96.0\,/\,96.7\,/\,\textbf{97.4} \\
& Places365 &       42.0\,/\,34.1\,/\,25.8\,/\,\textbf{25.7}        & 91.4\,/\,92.6\,/\,94.6\,/\,\textbf{94.6} \\
& CIFAR-100 &       53.7\,/\,45.2\,/\,45.1\,/\,\textbf{44.7}        & 88.3\,/\,89.9\,/\,90.7\,/\,\textbf{90.8} \\
& TIN (c) &         8.1\,/\,20.9\,/\,8.1\,/\,\textbf{4.2}           & 98.5\,/\,96.2\,/\,98.5\,/\,\textbf{99.1} \\
& LSUN (r) &        3.0\,/\,13.4\,/\,2.7\,/\,\textbf{1.3}           & 99.3\,/\,97.4\,/\,99.3\,/\,\textbf{99.7} \\
& TIN (r) &         6.2\,/\,24.0\,/\,8.6\,/\,\textbf{4.5}           & 98.8\,/\,95.6\,/\,98.3\,/\,\textbf{99.1} \\
& iSUN &            2.8\,/\,16.1\,/\,2.7\,/\,\textbf{1.5}           & 99.3\,/\,97.0\,/\,99.3\,/\,\textbf{99.6} \\
\cline{2-4}
& \textbf{average} & 17.7\,/\,21.8\,/\,13.7\,/\,\textbf{12.2}       & 96.4\,/\,95.6\,/\,97.2\,/\,\textbf{97.5} \\ 
\hline

\multirow{10}*{\rotatebox[origin=c]{90}{CIFAR-100}} 
& SVHN &            65.6\,/\,78.2\,/\,\textbf{36.6}\,/\,63.9        & 85.2\,/\,83.6\,/\,\textbf{94.0}\,/\,89.5 \\
& LSUN (c) &        35.3\,/\,46.2\,/\,\textbf{25.4}\,/\,31.1        & 93.3\,/\,90.4\,/\,\textbf{95.4}\,/\,94.6 \\
& Textures &        80.0\,/\,40.7\,/\,21.7\,/\,\textbf{17.9}        & 77.2\,/\,91.7\,/\,95.5\,/\,\textbf{96.3} \\
& Places365 &       79.5\,/\,\textbf{76.6}\,/\,81.4\,/\,83.3        & 76.8\,/\,\textbf{77.5}\,/\,76.4\,/\,75.7 \\
& CIFAR-10 &        \textbf{83.6}\,/\,84.1\,/\,87.1\,/\,90.5        & 71.2\,/\,\textbf{75.0}\,/\,70.5\,/\,69.3 \\
& TIN (c) &         63.1\,/\,51.0\,/\,25.9\,/\,\textbf{14.5}        & 87.1\,/\,90.1\,/\,95.3\,/\,\textbf{97.5} \\
& LSUN (r) &        75.6\,/\,56.7\,/\,22.9\,/\,\textbf{6.5}         & 85.2\,/\,88.6\,/\,95.7\,/\,\textbf{98.7} \\
& TIN (r) &         73.5\,/\,51.0\,/\,20.6\,/\,\textbf{7.9}         & 84.6\,/\,89.8\,/\,96.0\,/\,\textbf{98.5} \\
& iSUN &            78.6\,/\,57.0\,/\,24.7\,/\,\textbf{10.5}        & 83.8\,/\,88.7\,/\,95.2\,/\,\textbf{97.9} \\
\cline{2-4}
& \textbf{average} & 69.5\,/\,60.1\,/\,38.5\,/\,\textbf{36.2}       & 82.7\,/\,86.1\,/\,90.4\,/\,\textbf{90.9} \\ 
\hline
\end{tabular}
\caption{Comparison with retraining methods. $\uparrow$ ($\downarrow$) indicates larger (smaller) values are better. \textbf{Bold} numbers are superior.}
\label{tab2}
\end{table*}

\subsection{Results and Discussions}
\subsubsection{Main Results.}
The main results are reported in Table \ref{tab1} and Table \ref{tab2}. For fair evaluation, we compare the proposed methods with competitive OOD detection algorithms which \textbf{do not rely on auxiliary OOD training data}. In Table \ref{tab1}, we show the performance of our method and other post-hoc methods based on discriminative models, including MSP \cite{hendrycks2016baseline}, ODIN \cite{liang2017enhancing}, Mahalanobis \cite{Lee2018ASU}, and Energy \cite{liu2020energy}. Over a total of 18 combinations of ID and OOD datasets, the proposed READ-MD algorithm outperforms the previous competing methods in 12 of them and gives second highest results on 2 of them.\footnote{This is based on the FPR@95TPR value; AUROC result is comparable.} Moreover, we show that using READ-MD reduces the average FPR@95TPR by \textbf{9.8\%} compared to the second best Energy score when CIFAR-10 is ID, and \textbf{4.0\%} compared to the second best Mahalanobis score when CIFAR-100 is ID. Without pre-training constraint, we present comparison results of the proposed READ-ED with three variants of G-OIDN \cite{hsu2020generalized} in Table \ref{tab2}, i.e., G-ODIN-I, G-ODIN-C, and G-ODIN-E. Our method reduces the average FPR@95TPR by \textbf{1.5\%} on ID CIFAR-10 compared to G-ODIN. The improvement is enlarged to \textbf{2.3\%} on complex ID CIFAR-100. In particular, both of our methods decrease the FPR@95TPR metric for hard OODs by a large margin. It is worth noting that retraining classifier slightly deteriorate the classification performance, from 94.85\% to 94.62\% for CIFAR-10 and 75.83\% to 75.08\% for CIFAR-100. 

\subsubsection{Combination Study.}
To investigate how the performance of OOD detection changes when combining $Score_{\mathrm{cla}}$ and $Score_{\mathrm{rec}}$, we present detailed results for separated and aggregated OOD score in Table \ref{tab3}. Empirically, the combination brings lower FPR@95TPR and higher AUROC for most OOD datasets across our methods. The rationale is that the two scores represent the discrepancy of ID and OOD data from different perspectives and achieve an effect of complementation.

\begin{table}[h]
\begin{tabular}{>{\centering\arraybackslash}p{0.1\textwidth}>{\centering\arraybackslash}p{0.15\textwidth}>{\centering\arraybackslash}p{0.15\textwidth}}
\hline
\multicolumn{3}{c}{$-Score_{\mathrm{cla}}$\,/\,$-Score_{\mathrm{rec}}$\,/\,$-(Score_{\mathrm{cla}} + Score_{\mathrm{rec}})$} \\
\multirow{1}*{\textbf{Method}} & \textbf{FPR@95TPR} $\downarrow$ & \textbf{AUROC} $\uparrow$ \\

\hline
READ-MD & 46.3\,/\,55.3\,/\,\textbf{37.6}        & 90.2\,/\,75.4\,/\,\textbf{90.8} \\ 
\hline

READ-ED & 13.7\,/\,78.7\,/\,\textbf{12.4}      &  97.2\,/\,59.1\,/\,\textbf{97.5} \\
\hline
\end{tabular}
\caption{OOD detection results for combination study. $\uparrow$ ($\downarrow$) indicates larger (smaller) values are better. The results are averaged on nine OOD test datasets. \textbf{Bold} numbers are superior results.}
\label{tab3}
\end{table}

\subsubsection{Ablation Study.}
Table \ref{tab4} validates the contributions of reconstruction error adjustment coefficient and input perturbation techniques. We report the average detection performance over 9 OOD datasets when CIFAR-10 is used as ID. After gradually applying techniques to our score function, one can note that reconstruction error adjustment decreases FPR@95TPR by \textbf{5.6\%}  for READ-MD. We do not present the ablation study result for READ-ED since the perturbation magnitude $\epsilon$ searched by ID and synthetic OOD data equals to 0. It is clear that the overconfidence problem for easy OOD of \texttt{AE} is alleviated. Therefore, the proposed adjustment coefficient is an indispensable part that strengthens our methods.

\begin{table}[t]
\begin{tabular}{>{\centering\arraybackslash}p{0.08\textwidth}>{\centering\arraybackslash}p{0.02\textwidth}>{\centering\arraybackslash}p{0.02\textwidth}>{\centering\arraybackslash}p{0.14\textwidth}>{\centering\arraybackslash}p{0.1\textwidth}}
\hline
\textbf{Method} & \textbf{Adj} &\textbf{Pert} & \textbf{FPR@95TPR} $\downarrow$ & \textbf{AUROC} $\uparrow$ \\
\hline
\multirow{4}*{READ-MD} 
& - & - & 37.6 & 90.8 \\
& - & \checkmark &  29.9 &  92.7 \\
& \checkmark & - &  33.3 &  92.3 \\
& \checkmark & \checkmark  & \textbf{28.5} & \textbf{93.4} \\

\hline
\end{tabular}
\caption{OOD detection results for ablation study. $\uparrow$ ($\downarrow$) indicates larger (smaller) values are better. \textbf{Bold} numbers are superior results. Adj and Pert mean adjustment and perturbation respectively.}
\label{tab4}
\end{table}



\section{Conclusion}
In this work, we propose READ for out-of-distribution detection. The key idea is to unify distance to the closest class and reconstruction error in the latent space of classifier. We show that the combination of transformed reconstruction error exhibits superior detection performance. Against the overconfidence issue of autoencoder, we adjust the transformed reconstruction error with an image complexity based coefficient. As a result, the variants of READ, namely READ-MD and READ-ED, both achieve state-of-the-art performance in the corresponding scenario. Extensive ablations provide further understandings of our methods. We hope future work will pay more attention to mining and combining the inconsistencies of ID and OOD from different models.

\section*{Acknowledgements}
This paper is supported by the National Natural Science Foundation of China (Grant No. 62192783, U1811462), the
Collaborative Innovation Center of Novel Software Technology and Industrialization at Nanjing University.

%
%
%
\bibliography{citations}

\end{document}